\documentclass[accepted]{uai2021} 

\usepackage[american]{babel}
\usepackage{url}            
\usepackage{booktabs}       
\usepackage{nicefrac}       
\usepackage{microtype}      
\usepackage{bm}
\usepackage{bbm}
\usepackage{amsmath}
\usepackage{amssymb}
\usepackage{amsthm}

\usepackage{natbib} 
\usepackage{mathtools} 
\usepackage{siunitx} 
\usepackage{tikz} 

\newtheorem{theorem}{Theorem}

\usepackage{comment}
\usepackage{caption}
\usepackage{subcaption}
\usepackage{algorithm}
\usepackage{algorithmic}

\title{Generative Archimedean Copulas}

\author[1]{\href{mailto:Yuting Ng <yuting.ng@duke.edu>?Subject=Generative Archimedean Copulas}{Yuting Ng}{}} 
\author[2]{Ali Hasan}
\author[1]{Khalil Elkhalil}
\author[1]{Vahid Tarokh}
\affil[1]{%
    Department of Electrical and Computer Engineering\\
    Duke University
}
\affil[2]{%
    Department of Biomedical Engineering\\
    Duke University
}

\begin{document}
\maketitle

\begin{abstract}
We propose a new generative modeling technique for learning multidimensional cumulative distribution functions (CDFs) in the form of copulas. Specifically, we consider certain classes of copulas known as Archimedean and hierarchical Archimedean copulas, popular for their parsimonious representation and ability to model different tail dependencies. We consider their representation as mixture models with Laplace transforms of latent random variables from generative neural networks. This alternative representation allows for computational efficiencies and easy sampling, especially in high dimensions. We describe multiple methods for optimizing the network parameters. Finally, we present empirical results that demonstrate the efficacy of our proposed method in learning multidimensional CDFs and its computational efficiency compared to existing methods.
\end{abstract}

\section{Introduction}\label{sec:intro}

Copulas are a special class of cumulative distribution functions (CDFs) that model the dependencies between multiple random variables in isolation from their marginals \citep{nelsen2007introduction, joe2014dependence}. Copulas have found applications in many areas, including hydrology~\citep{genest2007everything} and finance~\citep{cherubini2004finance}. In finance, for example, more expressive modeling using copulas of the joint distribution of two stocks can result in more pairs trading opportunities~\citep{botha2013_coppairstrading,liew2013_coppairstrading}.

In machine learning, copulas have been used to create new distributions, increasing the flexibility of modeling multivariate dependencies~\citep{ghahramani2010_copulaproc, elidan2010_copulabayesian, huang2011_cdn, tagasovska2019vae, wiese2019copulaflows, kamthe2021copulaflows, chilinski2020_neuralLikelihoods}. The utility of copulas can be attributed to their powerful representation capabilities, ease of use and intuitive decomposition into marginals and a dependence function. However, many challenges related to parameterization and estimation are still unsolved. 

A particularly useful class of copulas are known as Archimedean copulas, which endow a specific structure for representing the dependence function in terms of a one-dimensional \emph{generator} function. 
Most work involving Archimedean copulas consider different parameterizations for this generator function. 
Parameterizations of the generator function have generally been limited to simple forms, since complicated generator functions lead to difficulties in computing the copula density, a necessary component for maximum likelihood estimation. 
\citet{ling2020_ACNet} proposed parameterizing the generator function as a neural network, but ran into computational difficulties for dimensions greater than 5. 
\textcolor{black}{We consider an alternative construction based on a mixture representation with latent random variables, first proposed by~\citet{marshallolkin1988sampling}, wherein we parameterize a latent distribution, whose Laplace transform acts as the generator function, with a generative neural network.}  Depending on the application, this latent variable is sometimes known as a \emph{resilience} or \emph{frailty} parameter. Using this construction, we can scale computations to higher dimensions and bypass numerical issues involved with automatic differentiation. 

Employing the Laplace transform to a learned latent model also provides important benefits beyond computational efficiency and numerical stability. When sampling from the copula using established approaches~\citep{marshallolkin1988sampling, mcneil2008sampling, hering2010_levy}, knowledge of the latent distribution is necessary. Parameterizing the latent distribution with a generative neural network allows for efficient sampling after training. 

Archimedean copulas can also be extended to the so-called hierarchical (or nested) Archimedean copulas, where multiple generators are used in conjunction to increase expressiveness of the model~\citep{joe1997multivariate}. 
This architecture mitigates a central deficiency of vanilla Archimedean copulas — the assumed symmetry in the dependence structure. 
We use a construction based on Lévy subordinators, i.e. non-decreasing Lévy processes such as the compound Poisson process, first proposed by~\citet{hering2010_levy}, and parameterize the Lévy subordinators using generative neural networks. 
We also use Laplace transforms as in the vanilla Archimedean copula to obtain the generator functions and subsequently recover a richer class of copulas. 

\subsection*{Related Work}
Our part of work on Archimedean copulas is related to~\citep{ling2020_ACNet}, where a neural network is proposed to represent the generator function of an Archimedean copula. We propose instead a generative neural network to represent the latent random variable, whose Laplace transform gives the generator function of the Archimedean copula. We then approximate the Laplace transform with the empirical Laplace transform using samples from the generative neural network. We note that there exist prior work that replace the Laplace transform with the empirical Laplace transform, such as those on the estimation of compound Poisson processes and distribution goodness-of-fit tests. These can be found in~\citep{csorgo1990empiricallaplace, henze2012gofempiricallaplace}, but do not consider/employ neural networks.

Existing semiparametric methods for Archimedean copulas are mainly concentrated on two dimensional cases, and their efficacy in higher dimensions remains unclear~\citep{lobato2011semiparametric, hoyos2020semiparametric}. Other work on the mixture representation with a latent random variable is limited to cases of known distributions that can be sampled and for which the Laplace transform can be calculated, since it is often challenging to find and sample from a distribution corresponding to arbitrary Laplace transforms~\citep{mcneil2008sampling, hofert2008sampling}.

Our part of work on hierachical Archimedean copulas is inspired by~\citep{hering2010_levy} who recognized that sufficient nesting conditions of hierarchical Archimedean copulas may be satisfied using Lévy subordinators. We then let the increments associated with the Lévy measure of the Lévy subordinator be the output of a generative neural network, and compute its integral in the Laplace exponent as an expectation with samples from the generative neural network. Related work parameterizing the Lévy measure with a neural network can be found in~\citep{xu2020_levy}, \textcolor{black}{but the integral is approximated as a Riemann sum, and} it does not relate to hierarchical Archimedean copulas. 

Other related works combine one-parameter families of Archimedean copulas, usually in a homogeneous manner, where all components are from the same family. It is challenging to combine Archimedean copulas from different families due to the nesting conditions. For example, the Clayton and Gumbel copulas are not compatible for nesting~\citep{mcneil2008sampling}. Thus, related works on heterogeneous Archimedean copulas have resulted in limited combinations of different families~\citep{mcneil2008sampling, hofert2008sampling, savu2010hac, okhrin2014hac, gorecki2017hac}.

\subsection*{Main Contributions}
First, we propose to use a generative neural network to represent the latent random variable, whose Laplace transform provides the generator function of an Archimedean copula. This allows approximation of the Laplace transform with its empirical version through samples from the generative neural network. Computing higher-order derivatives using the properties of the empirical Laplace transform additionally allows scalability to higher-dimensional data. Second, we extend this concept to modeling hierarchical Archimedean copulas with Lévy subordinators. We represent the Lévy measure of a Lévy subordinator with a generative neural network and compute its Laplace exponent using samples from the generative neural network. We then propose three methods for training: maximum likelihood with the copula density, goodness-of-fit with the Cramér-von Mises statistic, and adversarial training by minimizing a divergence between true samples from data and fake samples from the copula. Finally, we adapt existing Marshall-Olkin type efficient sampling algorithms to our parameterization with generative neural networks. The source code for this paper may be found at \url{https://github.com/yutingng/gen-AC}.

\subsection*{Outline}

Section~\ref{sec:background} provides the mathematical background on copulas, Archimedean copulas and hierarchical Archimedean copulas. Section~\ref{sec:generativeAC} discusses modeling, sampling and training generative Archimedean copulas. Section~\ref{sec:generativeHAC} extends the construction to hierarchical Archimedean copulas. Section~\ref{sec:experiments} shows our experiment results on learning known Archimedean and hierarchical Archimedean copulas that have different tail dependencies. We also compare its flexibility in fitting real-world data to commonly-used one-parameter families. In addition, we show its computational efficiency and sampling in higher-dimensions. Finally, we conclude the paper in Section~\ref{sec:conclusion}.

\section{Background}\label{sec:background}

We begin by describing the necessary background on copulas. A \emph{copula} is a multivariate cumulative distribution function (CDF) where all univariate margins are uniform, i.e. it is the CDF of a vector of dependent uniform random variables. Multidimensional dependence modeling with copulas is based on a theorem due to~\citet{sklar1959} which gives a general representation of a multivariate CDF as a composition of its univariate margins and a copula.

\begin{theorem}[Sklar's theorem]
For a $d-$variate cumulative distribution function $F$, with $j$th univariate margin $F_j$, and $j^{th}$ quantile function $F_j^{-1}$, the copula associated with $F$ is a cumulative distribution function $C:[0,1]^d\to [0,1]$ with $\text{U}(0,1)$ margins satisfying:
\begin{align}
    F(\mathbf{x}) &=C(F_1(x_1),\cdots,F_d(x_d)),\; \mathbf{x}\in\mathbb{R}^d,\\
    C(\mathbf{u}) &= F(F_1^{-1}(u_1),\cdots,F_d^{-1}(u_d)),\; \mathbf{u}\in[0,1]^d.
\end{align}
In addition, if $F$ is continuous, then $C$ is unique.
\end{theorem}

Moreover, due to Sklar's theorem, every CDF endows such a decomposition.  Thus, copulas allow characterization of the multivariate dependence between the random variables $X_1,\cdots,X_d$ separately from their univariate margins $F_1,\cdots,F_d$~\citep{nelsen2007introduction,joe2014dependence}. 

\subsection{Archimedean Copulas}

An important class of copulas are the Archimedean copulas, due to their ease of construction and ability to represent different tail dependencies. An Archimedean copula is defined as:
\begin{equation}\label{eq:ac_distribution}
    C(\mathbf{u}) = \varphi\left(\varphi^{-1}(u_1)+\cdots+\varphi^{-1}(u_d)\right),
\end{equation}
with density:
\begin{align}\label{eq:ac_density}
c(\mathbf{u}) &= \frac{\partial^{d}C(u_1,\cdots,u_d)}{\partial u_1\cdots\partial u_d} \\\nonumber
&= \frac{\varphi^{(d)}(\varphi^{-1}(u_1)+\cdots+\varphi^{-1}(u_d))}{\prod_{i=1}^d\varphi'(\varphi^{-1}(u_i))}.
\end{align}

For the above expression to be a valid copula for all $d$, the one-dimensional function $\varphi : [0,\infty)\to[0,1]$, known as the \emph{generator} of the Archimedean copula must satisfy:
\begin{itemize}
    \item $\varphi(0)=1, \; \varphi(\infty)=0$,
    \item $\varphi$ is \emph{completely monotone}, \\i.e. $(-1)^k \varphi^{(k)} \geq 0$ for all $k \in \{0,1,2,\cdots\}$.
\end{itemize} The criteria that $\varphi$ is completely monotone, i.e. its derivatives change signs, guarantees positiveness of the copula density~\citep{kimberling1974}. The class of completely monotone $\varphi$ coincides with the class of Laplace-Stieltjes transforms (henceforth simply Laplace transforms) of a positive random variable~\citep{bernstein1929fonctions, widder1941laplace}. 

 \begin{theorem}
 [\citet{bernstein1929fonctions} and~\citet{widder1941laplace}] $\varphi$ is completely monotone and $\varphi(0) = 1$ if and only if $\varphi$ is the Laplace transform of a positive random variable, 
\begin{equation}
    \varphi(x) = \int_0^\infty e^{-xs}dF_M(s),
\end{equation}
where $M>0$ is a positive random variable with Laplace transform $\varphi$.
 \end{theorem}

Conversely, a probabilistic construction of the Archimedean copula as a mixture model, with the variables being conditionally independent given a positive latent random variable, leads to the Laplace transform representation for $\varphi$. For a given $d$, $\varphi$ may come from a broader class of functions than Laplace transforms~\citep{mcneil2009dmonotone}. However, if $\varphi$ is not a Laplace transform, the simple mixture representation fails~\citep{marshallolkin1988sampling}. In the mixture representation, the latent variable, depending on its application, is known as a \emph{resilience} or \emph{frailty} parameter~\citep{marshallolkin1988sampling, joe1997multivariate}. Common Archimedean copulas such as the Ali-Mikhail-Haq, Clayton, Frank, Gumbel and Joe copulas can be respectively derived from geometric, gamma, logarithmic, stable, and Sibuya latent distributions. The mixture representation also leads to efficient sampling algorithms~\citep{marshallolkin1988sampling, mcneil2008sampling}.

We restate the probabilistic construction and sampling algorithm in the supplementary material.

\subsection{Hierarchical Archimedean Copulas}

While Archimedean copulas have been widely employed, the functional symmetry of the Archimedean copula implies exchangeability of the underlying dependence structure, which is sometimes not realistic. Hierarchical (or nested) Archimedean copulas are popular for overcoming this drawback~\citep{joe1997multivariate}. 

\begin{figure}[!h]
    \centering
    \includegraphics[scale=0.9, trim = 0cm 0cm 0cm 0cm, clip]{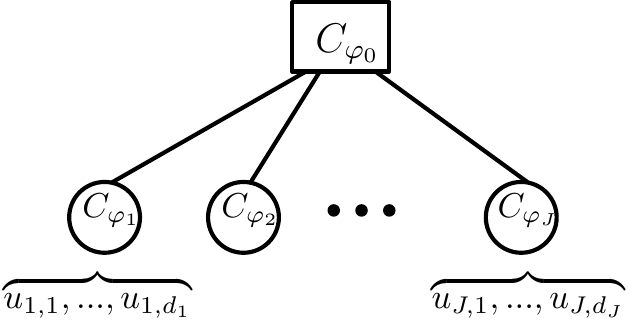}
    \caption{Hierarchical Archimedean copula with $J$ nested, possibly hierarchical, Archimedean copulas.}
    \label{fig:hierarchical_AC}
\end{figure}
In this case, the copula can be written as:
\begin{equation}\label{eq:hierarchical_example}
    C(\mathbf{u}) = C_{\varphi_0}(C_{\varphi_1}(\mathbf{u}_{1}),\cdots,C_{\varphi_J}(\mathbf{u}_{J})),
\end{equation}
where $C_{\varphi_j}$ are nested, possibly hierarchical, Archimedean copulas with generators $\varphi_j$, inputs $\mathbf{u}_j=[u_{j,1}, \cdots, u_{j,d_j}]$, $j\in\{1,\cdots,J\}$, and $\mathbf{u} = [\mathbf{u}_1, \cdots, \mathbf{u}_J]$, $d_1 + \cdots + d_J=d$.

For the above expression to be a valid copula, additional sufficient nesting conditions, derived from the nested mixture representation, first given in~\citep{joe1997multivariate} and restated in~\citep{mcneil2008sampling} for nesting to arbitrary depth are:
\begin{itemize}
    \item $\varphi_j$ for all $j\in\{0,1,\cdots,J\}$ are completely monotone,
    \item $(\varphi_0^{-1}\circ\varphi_j)'$ for $j\in\{1,\cdots,J\}$ are completely monotone.
\end{itemize}

The criteria that $(\varphi_0^{-1}\circ\varphi_j)'$ are completely monotone come from the composition of an \emph{outer generator} $\varphi_0$ and an \emph{inner generator} $\varphi_j$ to produce a completely monotone Laplace transform \emph{nested generator} of the form $e^{-M\varphi_0^{-1}\circ\varphi_j}$, where $M$ is distributed with Laplace transform $\varphi_0$~\citep{joe1997multivariate, mcneil2008sampling}. This criteria is addressed in~\citep{hering2010_levy} using Lévy subordinators, i.e. non-decreasing Lévy processes such as the compound Poisson process, by recognizing that the Laplace transform of Lévy subordinators at a given `time' $t\geq0$ have the form $e^{-t\psi_j}$, where the Laplace exponent $\psi_j$ has completely monotone derivative. Conversely, a probabilistic construction by combining Lévy subordinators evaluated at common `time' $t=M$, leads to a well-defined hierarchical Archimedean copula with an efficient sampling algorithm~\citep{hering2010_levy}.
We restate the probabilistic construction from~\citep{hering2010_levy} in the supplementary material.

Thus for a given \emph{outer generator} $\varphi_0$, a \emph{compatible inner generator} $\varphi_j$ can be modeled as a composition of the outer generator and the Laplace exponent $\psi_j$ of a Lévy subordinator:
\begin{equation}\label{eq:inner_generator}
    \varphi_j(x) = (\varphi_0 \circ \psi_j)(x),
\end{equation}
where the Laplace exponent $\psi_j:[0,\infty)\to[0,\infty)$ of a Lévy subordinator has a convenient representation with drift $\mu_j\geq0$ and Lévy measure $\nu_j$ on $(0,\infty)$ due to the Lévy-Khintchine theorem~\citep{ken1999levy}:
\begin{equation}\label{eq:Laplace_exponent}
    \psi_j(x) = \mu_j x + \int_0^\infty (1-e^{-xs}) \nu_j(ds).
\end{equation} 

A popular Lévy subordinator is the compound Poisson process with drift $\mu_j\geq0$, jump intensity $\beta_j>0$ and jump size distribution determined by its Laplace transform $\varphi_{M_j}$. In this case, the Laplace exponent has the following expression:
\begin{align}\label{eq:compound_poisson}
    \psi_j(x) &= \mu_j x + \beta_j (1-\varphi_{M_j}(x))\\\nonumber
            &= \mu_j x + \beta_j (1-\int_0^\infty e^{-xs}dF_{M_j}(s)),
\end{align}
where $M_j>0$ is a positive random variable with Laplace transform $\varphi_{M_j}$ characterizing the jump sizes of the compound Poisson process.

In addition, we choose $\mu_j>0$ to satisfy the condition $\varphi_j(\infty)=(\varphi_0 \circ \psi_j)(\infty)=0$ such that $\varphi_j$ is a valid generator of an Archimedean copula. 

\section{Generative Archimedean Copulas}\label{sec:generativeAC}

Motivated by the probabilistic construction of the Archimedean copula, we propose to learn the distribution of the positive latent variable by approximating its Laplace transform using samples from a generative neural network.

\subsection{Modeling the Latent Variable with a Generative Neural Network}~\label{sec:generativeM}

We let $M$ be the output of a generative neural network such that samples $M\sim F_M$ are computed as $M = G(\epsilon; \theta)$, where $G(. ; \theta)$ represents the generative neural network with parameters $\theta$ and $\epsilon$ is a source of randomness. Unlike the modeling of monotone functions with neural networks~\citep{chilinski2020_neuralLikelihoods}, there is no restriction on the weights and intermediate activations of $G(. ; \theta)$. In this preliminary work, the network architecture is a multilayer perceptron. To guarantee that $M$ is a positive random variable, we use $\exp(.)$ as the output activation.

We then approximate the Laplace transform with its empirical version using $L$ samples of $M$ from $G(.;\theta)$ as:
\begin{equation}\label{eq:Laplace}
    \varphi(x) = \int_0^\infty e^{-xs}dF_M(s) = \mathbb{E}_M[e^{-Mx}] \approx \frac{1}{L}\sum_{l=1}^L e^{-M_lx}.
\end{equation}

Derivatives of the Laplace transform are similarly approximated with their empirical version as:
\begin{equation}\label{eq:Laplace_derivative}
    \varphi^{(k)}(x)=\mathbb{E}_M[(-M)^ke^{-Mx}] \approx \frac{1}{L}\sum_{l=1}^L (-M_l)^ke^{-M_lx}.
\end{equation}

Subsequently, we replace instances of $\varphi$ and $\varphi^{(k)}$ in the copula distribution, density and sampling algorithm with their sample approximations computed as in \eqref{eq:Laplace} and \eqref{eq:Laplace_derivative}.

\subsection{Generating Samples from the Archimedean Copula}\label{sec:sampling}

\begin{figure}[!h]
    \centering
    \includegraphics[scale=0.8, trim = 0cm 2.5cm 0cm 0cm, clip]{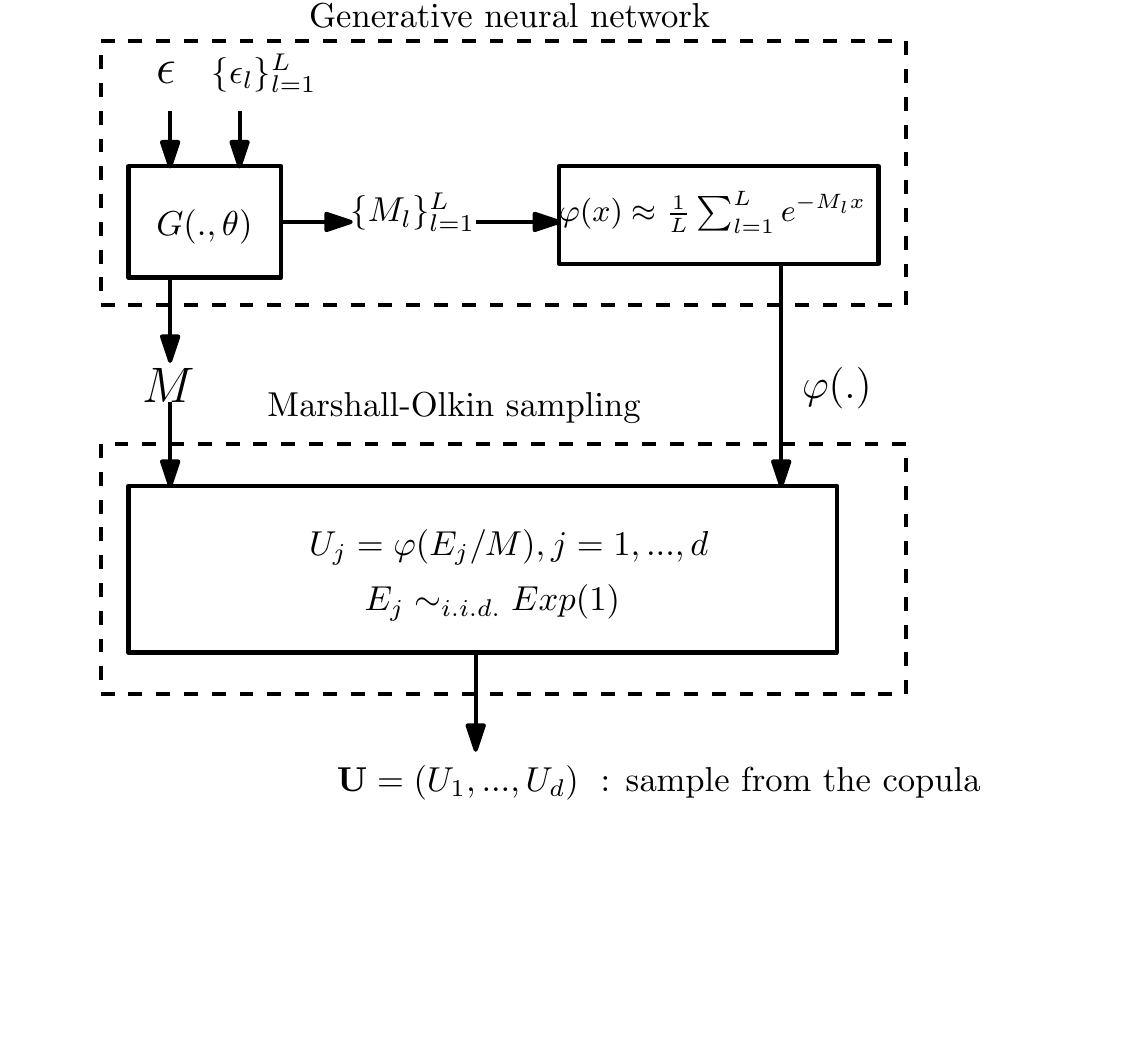}
    \caption{Sampling Archimedean copulas using generative neural networks and Marshall-Olkin type sampling.}
    \label{fig:sampling}
\end{figure}

We modify existing Marshall-Olkin type sampling algorithms~\citep{marshallolkin1988sampling, mcneil2008sampling} to our parameterization with generative neural networks, as detailed in Algorithm \ref{alg:sampling_gac} and~\figurename~\ref{fig:sampling}, on the next page.

This sampling method is efficient as it only requires sampling unit exponential random variables $E_j\sim\text{Exp}(1),\;j\in\{1,\cdots,d\}$ and a latent random variable $M= G(\epsilon;\theta)$. In addition, unlike the conditional sampling method, this sampling method does not require differentiation of the copula distribution to get the conditional distribution and does not require inversion of the conditional distribution.

\begin{algorithm}
\caption{Sampling Generative Archimedean Copulas}\label{alg:sampling_gac}
\textbf{Input:} $G(.;\theta)$,
	\begin{algorithmic}[1]
	\STATE Sample $M$ as $M = G(\epsilon;\theta).$ 
	\STATE Sample i.i.d. $E_j\sim\text{Exp}(1), j\in\{1,\cdots,d\}$.
	\STATE Approximate $\varphi$ with samples $\{M_l \}_{l=1}^L$, where $M_l = G(\epsilon_l;\theta)$, as in~\eqref{eq:Laplace}.
	\STATE Compute $\mathbf{U}$ where $U_j=\varphi(E_j/M), j\in\{1,\cdots,d\}$.
	\end{algorithmic} 
\textbf{Output:} $\mathbf{U}$.
\end{algorithm}

\subsection{Training Methods}
An important consideration when modeling CDFs is the optimization procedure for fitting the model to data. 
We describe multiple methods for fitting the model to data with various performance and efficiency trade-offs. 

\subsubsection{Training with Maximum Likelihood}

We consider training through maximum likelihood by minimizing the negative log likelihood with backpropagation gradient descent on the model parameters, similar to the proposal in~\citep{ling2020_ACNet}. However, since the copula models the CDF, differentiation is required to obtain the copula density. Unlike~\citep{ling2020_ACNet} that computes the copula density from the copula distribution using automatic differentiation, we compute the copula density from its analytical expression in~\eqref{eq:ac_density} using the properties of the Laplace transform for computing higher-order derivatives in~\eqref{eq:Laplace_derivative}. For increasing dimensions, computing higher-order derivatives using the Laplace transform representation instead of automatic differentiation leads to a significant speed up in computation.

For the computation of $\varphi^{-1}$ and its derivative with respect to model parameters, we borrow the method in~\citep{ling2020_ACNet}. The inverse is computed using Newton's root-finding method. The derivatives are computed from the derivatives of $\varphi$ then supplemented to backpropagation.

\subsubsection{Training with Goodness-of-Fit}

To circumvent computing the copula density, the model may also be fitted to data via minimum distance criterions used in goodness-of-fit tests~\citep{genest2009gof}. Though not statistically efficient compared to maximum likelihood estimation,
minimum distance estimation is significantly less computationally intensive.

We consider the Cramér-von Mises statistic~\citep{cramer1928cvm} to measure a discrepancy between the model copula $C_\theta$ and the empirical copula $C_N$:
\begin{equation}\label{eq:cvm}
   S_N =\frac{1}{N}\sum_{i=1}^N (C_\theta(\mathbf{u}_i) - C_N(\mathbf{u}_i))^2,
\end{equation}

where $\mathbf{u}_i$ is an observation of the margins, $N$ is the number of observations and $C_N$ is the empirical copula given by:
\begin{equation}
    C_N(\mathbf{u})=\frac{1}{N}\sum_{i=1}^N \mathbbm{1}\{u_{i,1}\leq u_{1},\cdots,u_{i,d}\leq u_{d}\}.
\end{equation}

\subsubsection{Adversarial Training with Samples}

An alternative way to train the model is by minimizing a divergence between true samples from data and fake samples from the model copula, similar to generative adversarial networks (GANs)~\citep{goodfellow2014gan}. In this case, we solve the minimax problem in GANs where the generating network must satisfy an Archimedean copula. This is another method that allows training without computing the copula density. 

We create a discriminative neural network $D(. ; \phi)$ with parameters $\phi$ and $\text{sigmoid}(.)$ output activation to distinguish between true samples from data and fake samples from the copula. We then minimize the Jensen-Shannon loss between true and fake samples as in~\citep{goodfellow2014gan}:
\begin{equation}
    \min_{\theta}\max_{\phi}\mathbf{E}_{U\sim\text{data}}[\log(D(U;\phi))]+\mathbf{E}_{\tilde{U}\sim C}[\log(1-D(\tilde{U};\phi))],
\end{equation}
where $\tilde{U}\sim C$ is generated via the sampling method described in Algorithm~\ref{alg:sampling_gac} using the latent random variable represented as the output of the generative neural network $G(.;\theta)$ with parameters $\theta$ as discussed in Section~\ref{sec:generativeM}.

\section{Generative Hierarchical Archimedean Copulas}\label{sec:generativeHAC}

In the following, we extend the application of generative neural networks to hierarchical Archimedean copulas. We present our results for two levels of hierarchy, but our construction extends to nesting with more levels.

\subsection{Modeling the Laplace Exponent with a Generative Neural Network}

For a given outer generator $\varphi_0$, the inner generator $\varphi_j, j\in\{1,\cdots,J\}$ is  obtained as the composition $\varphi_j=\varphi_0 \circ \psi_j$, as in~\eqref{eq:inner_generator}, where $\psi_j$ is the Laplace exponent of a compound Poisson process with Lévy-Khintchine representation, as in~\eqref{eq:compound_poisson}.  
We let the drift $\mu_j>0$ and the jump intensity $\beta_j>0$ be trainable parameters with $\exp(.)$ output activation. We let the jump size $M_j>0$ be the output of a generative neural network $G(.;\theta_j)$ with parameters $\theta_j$ and $\exp(.)$ output activation. We then compute the Laplace transform $\varphi_{M_j}$ and its derivatives $\varphi_{M_j}^{(k)}$ using samples from $G(.;\theta_j)$ as in~\eqref{eq:Laplace} and~\eqref{eq:Laplace_derivative}.
\subsection{Generating Samples from the Hierarchical Archimedean Copula}
We modify the Marshall-Olkin type algorithm given in~\citep{hering2010_levy} to work with our parameterization using generative neural networks. 
We first describe sampling of a compound Poisson process in Algorithm~\ref{alg:sampling_cpp}. We then describe sampling of a generative hierarchical Archimedean copula in Algorithm~\ref{alg:sampling_ghac}. A sample from the hierarchical Archimedean copula is obtained by combining compound Poisson processes evaluated at a common `time' $t=M$, where $M$ is the random variable with distribution given by the Laplace transform outer generator $\varphi_0$.

\begin{algorithm}
\caption{Sampling compound Poisson process with jump sizes parameterized by generative neural network}
\label{alg:sampling_cpp}
\textbf{Input:} $t, \mu_j, \beta_j, G(.;\theta_j)$, 
	\begin{algorithmic}[1]
	\STATE Sample $N_j(t)\sim\text{Pois}(\beta_j t)$, i.e. the number of jumps by time $t$ of a Poisson random variable with rate $\beta_j$. 
	\STATE Sample $N_j(t)$ samples of $M_j$ from $G(.;\theta_j)$.
	\STATE Compute $\Lambda_j(t)=\mu_j t + \sum_{i=1}^{N_j(t)}M_{j,i}$.
	\end{algorithmic} 
\textbf{Output:} $\Lambda_j(t)$.
\end{algorithm}

\begin{algorithm}
\caption{Sampling Generative Hierarchical Archimedean Copulas}\label{alg:sampling_ghac}
\textbf{Input:} $G(.;\theta_0),\{\mu_j, \beta_j, G(.;\theta_j)\}_{j=1}^J$,\hfill \hbox{\hskip 20pt}\\
	\begin{algorithmic}[1]
	\STATE Sample $t=M$ from $G(.;\theta_0)$.
	\STATE Approximate $\varphi_0$ with samples from $G(.;\theta_0)$, as in~\eqref{eq:Laplace}.
	\FOR{$j\in\{1,\cdots,J\}$}
	    \STATE Sample $\Lambda_j(M)$, the compound Poisson process at $t=M$, following Algorithm~\ref{alg:sampling_cpp}.
	    \STATE Approximate $\psi_j$ with samples from $G(.;\theta_j)$, as in~\eqref{eq:Laplace}.
	\ENDFOR\vskip 10pt
	\STATE Sample i.i.d. $E_{j,i}\sim\text{Exp}(1)$, $j\in\{1,\cdots,J\}$, $i\in\{1,\cdots,d_j\}$.
    \STATE Compute $\mathbf{U}=(U_{1,1},\cdots,U_{J,d_J})$ as $U_{j,i}=(\varphi_0 \circ \psi_j)(E_{j,i}/\Lambda_j(M)), j\in\{1,\cdots,J\}, i\in\{1,\cdots,d_J\}$.
	\end{algorithmic} 
\textbf{Output:} $\mathbf{U}$.
\end{algorithm}

\subsection{Training with Goodness-of-Fit and Maximum Likelihood}

We first fit the outer generator $\varphi_0$, fix it, then fit the inner generators $\varphi_j=\varphi_0\circ\psi_j$. Fixing the outer generator then optimizing the inner generator provides additional numerical stability during training. In our experiments, the outer generator was trained using minimium distance estimation with the empirical copula based Cramér-von Mises statistic in~\eqref{eq:cvm} and empirical copulas on $C_{\varphi_j}$. An alternative method may be to train the outer generator using a composite likelihood with bivariate margins since bivariate margins are Archimedean with generator given by the outer generator. The inner generators were trained using maximum likelihood estimation with copula densities $c_{\varphi_j}$ in~\eqref{eq:ac_density}. 

\section{Experiments}\label{sec:experiments}

\subsection{Generative Archimedean Copula}

\subsubsection{Learning Bivariate Copulas with Different Tail Dependencies and Fitting Real-World Data}\label{sec:ex_bivariate}

Following the experiment setup in~\citep{ling2020_ACNet}, we consider the Clayton, Frank, and Joe copulas, chosen for their different tail dependencies, and the following real-world data sets: Boston housing, Intel-Microsoft (INTC-MSFT) stocks and Google-Facebook (GOOG-FB) stocks. We applied the three training methods discussed earlier: maximum likelihood, goodness-of-fit and adversarial training. All training methods were implemented in PyTorch and converged within 10k epochs. Experiment details are given in the supplementary material.

The negative log-likelihoods from learning known copulas are reported in~\tablename~\ref{tab:nll_common}. We use the following shorthands `GT', `ACNet', `MLE', `CvM', 'GAN' to respectively denote ground truth, ACNet~\citep{ling2020_ACNet}, and generative Archimedean copulas trained with maximum likelihood, goodness-of-fit and adversarial training. The negative log-likelihoods from fitting real-world data are reported in~\tablename~\ref{tab:nll_realworld}, where the log-likelihood of the best-fit single parameter copula (chosen from Clayton, Frank, Joe and Gumbel, as in~\citep{ling2020_ACNet}), with shorthand `BF' is reported in place of the ground truth.  The proposed generative Archimedean copulas achieved comparable performance to ACNet in terms of log-likelihood scores. In addition, out of the three methods, training with maximum likelihood achieved the best results; however, its increased computation cost, due to computing derivatives and inverses, motivates the use of the proposed alternative losses.

\begin{table}[!h]
    \centering
    \caption{Negative log-likelihoods of learning known copulas}\label{tab:nll_common}
    \begin{tabular}{rlllll}
      \toprule 
      &\multicolumn{2}{c}{\bfseries Benchmark} & \multicolumn{3}{c}{\bfseries Generative AC}\\
      \bfseries Dataset &  GT &  ACNet &  MLE &  CvM & GAN\\
      \midrule 
      Clayton & -0.94 & -0.92 & -0.89 & -0.86 & -0.89\\
      Frank & -0.90 & -0.88 & -0.89 & -0.86 & -0.89\\
      Joe & -0.51 & -0.49 & -0.48 & -0.35 & -0.47\\
      \bottomrule 
    \end{tabular}\vskip 10pt
    \centering
    \caption{Negative log-likelihoods of fitting real-world data}\label{tab:nll_realworld}
    \begin{tabular}{rlllll}
      \toprule 
      &\multicolumn{2}{c}{\bfseries Benchmark} & \multicolumn{3}{c}{\bfseries Generative AC}\\
      \bfseries Dataset &  BF &  ACNet &  MLE &  CvM &  GAN\\
      \midrule 
      Boston & -0.30 & -0.27 & -0.29 & -0.30 & -0.28\\
      INTC-MSFT & -0.19 & -0.20 & -0.16 & -0.15 & -0.17\\
      GOOG-FB & -0.93 & -0.96 & -0.95 & -0.92 & -0.94\\
      \bottomrule 
    \end{tabular}
\end{table}

\begin{figure}[!h]
    \centering
    \begin{subfigure}[b]{0.5\textwidth}
        \centering
        \includegraphics[width=\textwidth]{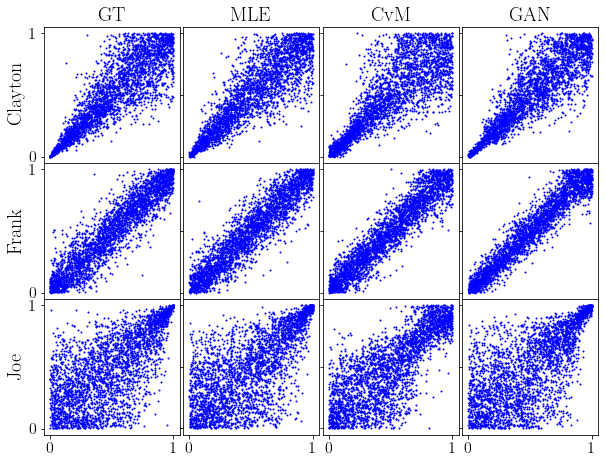}
        \caption{}\label{fig:samples_common}
    \end{subfigure}
    \vskip 10pt
    \begin{subfigure}[b]{0.5\textwidth}
        \centering
        \includegraphics[width=\textwidth]{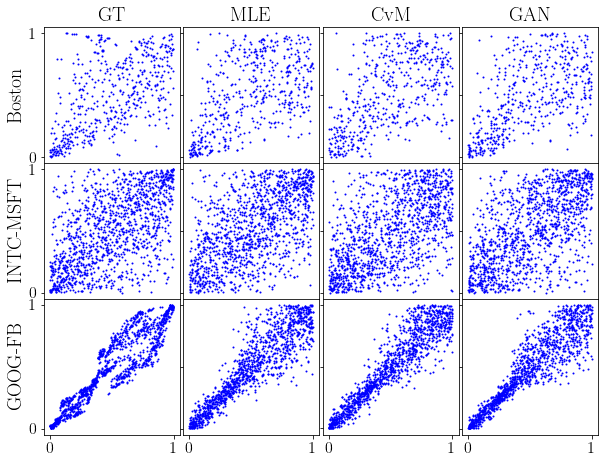}
        \caption{}\label{fig:samples_realworld}
    \end{subfigure}
    \caption{Samples from ground truth and learned copulas fitted with maximum likelihood, goodness-of-fit and adversarial training. In (a), the copulas are Clayton, Frank and Joe. In (b), the datasets are Boston housing, Intel-Microsoft stocks and Google-Facebook stocks.}\label{fig:samples}
\end{figure}

Samples from the learned copulas are compared to the ground truth in~\figurename~\ref{fig:samples}. We additionally note the differences in sampling time between our method and the conditional sampling method used in ACNet~\citep{ling2020_ACNet}. The time to generate 3000 samples using our method was on average $3.8 \times 10^{-2}$ seconds. In comparison, the conditional sampling method via automatic differentiation of the copula distribution followed by inversion of the conditional distribution, takes on average $1.98 \times 10^{+2}$ seconds, the difference on the order of 3 magnitudes.

\subsubsection{Learning Latent Distributions}

\begin{figure}[!h]
    \centering
    \includegraphics[width=\linewidth]{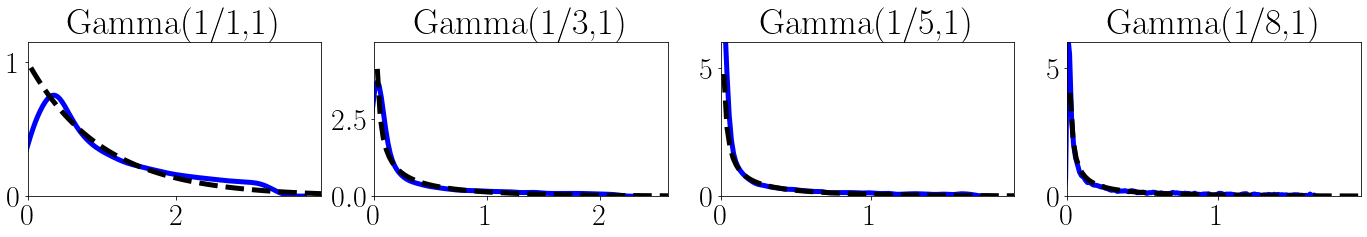}
    \caption{Gamma latent distributions of Clayton copulas with parameters 1, 3, 5, 8, learned in solid blue; ground truth in dashed black.} \label{fig:clayton_densities}
\end{figure}

The generative neural network was able to learn the latent Gamma distributions whose Laplace transforms give the generator functions of Clayton copulas. We show the learned latent distributions for Clayton copulas with parameters 1, 3, 5, 8 in~\figurename~\ref{fig:clayton_densities}.

\subsubsection{Learning Higher-Dimensional Copulas}

While ACNet faces numerical issues for dimensions $d\geq 5$ due to repeated automatic differentiation when computing the copula density~\citep{ling2020_ACNet}, the Laplace transform representation allows efficient computation of higher-order derivatives without automatic differentiation. 

In addition to the bivariate copulas in Section~\ref{sec:ex_bivariate}, we fitted Clayton, Frank and Joe copulas for 10 and 20 dimensions. The negative log-likelihoods are given in~\tablename~\ref{tab:nll_highd}. When compared to the ground truth negative log-likelihoods for 10-dimensional and 20-dimensional datasets, the learned negative log-likelihoods were off by 2\%. During our experiments, we could not obtain a reasonably trained ACNet for high dimensions due to the computational complexity.

\begin{table}[!h]
    \centering
    \caption{Negative log-likelihoods of learning higher-dimensional copulas}\label{tab:nll_highd}
    \begin{tabular}{rlllll}
      \toprule 
      &\multicolumn{2}{c}{\bfseries Ground Truth} & \multicolumn{2}{c}{\bfseries Generative AC}\\
      \bfseries Dataset &  10-dim &  20-dim &  10-dim &  20-dim\\
      \midrule 
      Clayton & -10.6 & -23.2 & -10.4 & -22.8\\
      Frank & -10.4 & -23.1 & -10.4 & -23.1\\
      Joe & -5.4 & -12.2 & -5.3 & -12.0\\
      \bottomrule 
    \end{tabular}\vskip 10pt
\end{table}

\begin{figure}[!h]
    \centering
    \includegraphics[width=0.7\linewidth]{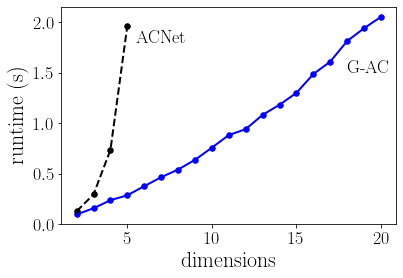}
    \caption{CPU runtimes for computing the likelihoods of 3000 samples from generative Archimedean copula in solid blue; ACNet~\citep{ling2020_ACNet} in dashed black.} \label{fig:density_time}
\end{figure}

Moreover, while the CPU runtimes of ACNet for computing the copula density increases exponentially with dimensions, the CPU runtimes of computing the copula density using the Laplace transform representation increases linearly with dimensions, as shown in~\figurename~\ref{fig:density_time}.

\subsection{Hierarchical Archimedean Copula}

\begin{figure}[!h]
    \centering
    \begin{subfigure}[b]{0.5\textwidth}
        \centering
        \includegraphics[width=\textwidth]{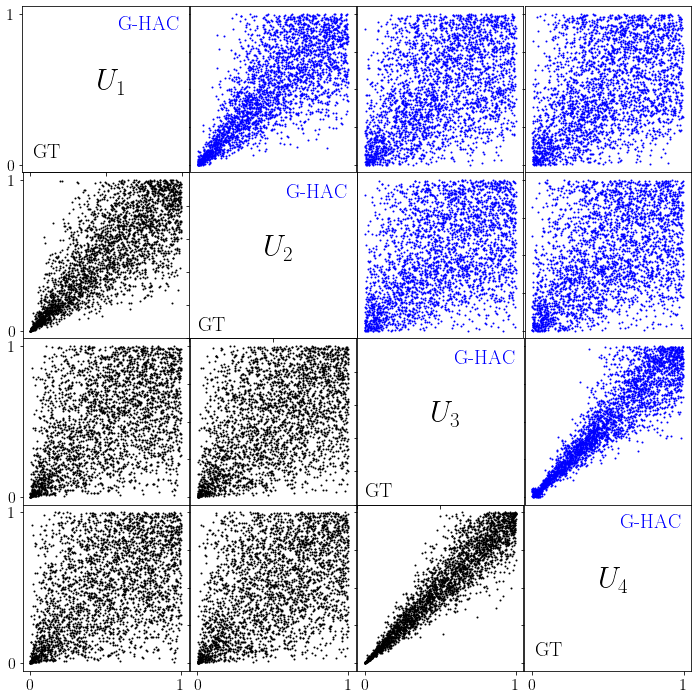}
        \caption{}
    \end{subfigure}
    \vskip 10pt
    \begin{subfigure}[b]{0.5\textwidth}
        \centering
        \includegraphics[width=\textwidth]{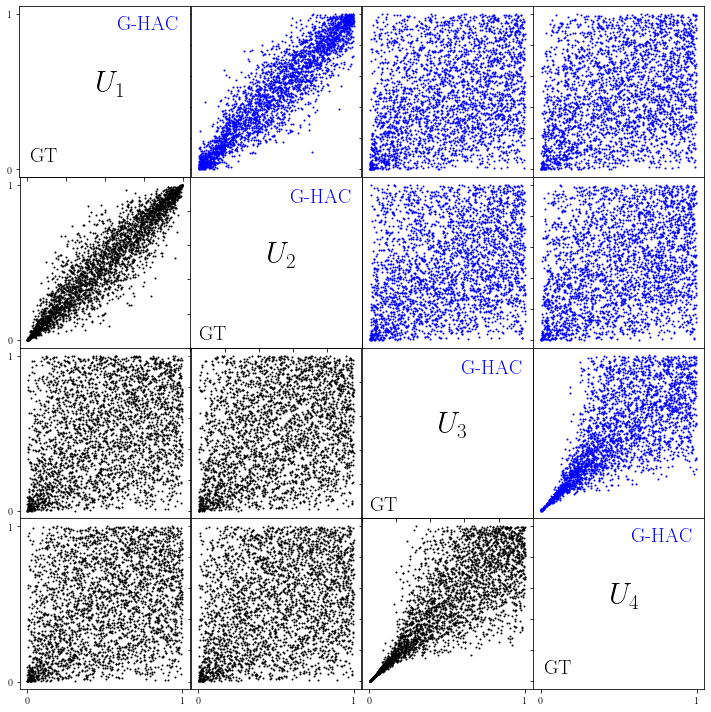}
        \caption{}
    \end{subfigure}
    \caption{Samples, displayed as mirrors on the diagonal, from generative hierarchical Archimedean copulas above in blue and from ground truth below in black. Each plot is a bivariate margin $(U_i,U_j)$. In (a), a homogeneous nested Clayton copula. In (b), a heterogeneous hierarchical Archimedean copula with a Clayton outer generator combined with inner generators `12' and `19', numbering following~\citep{nelsen2007introduction, gorecki2017hac}}. \label{fig:samples_4dim_clayton1}
\end{figure}

We demonstrate that our model can represent more complex dependence structures, beyond the exchangeability implied by the functional symmetry of Archimedean copulas, and learn hierarchical Archimedean copulas. 

We experiment with fitting a four-variate hierarchical Archimedean copula $C_{\varphi_0}(C_{\varphi_1}(u_1,u_2),C_{\varphi_2}(u_3,u_4))$. The ground truth was generated using the state-of-the-art HACopula Toolbox~\citep{gorecki2017hac}. Samples from the learned copulas are compared to the ground truth in~\figurename~\ref{fig:samples_4dim_clayton1}. In (a), $C_{\varphi_0}, C_{\varphi_1}, C_{\varphi_2}$ are Clayton copulas with parameters 1, 3, and 8. We let the outer generator be a generative Archimedean copula. In (b), $C_{\varphi_0}, C_{\varphi_1}, C_{\varphi_2}$ are Clayton, `12' and `19' with parameters 0.5, 3, and 1. Since our model is compatible with outer generators of other forms, we let the outer generator be a one-parameter Clayton copula instead of a generative Archimedean copula. 


\section{Conclusions}\label{sec:conclusion}

We modeled Archimedean and hierarchical Archimedean copulas with generative neural networks based on their probabilistic constructions as mixture and nested mixture models with latent random variables. We gave efficient sampling algorithms for sampling from the generative Archimedean and hierarchical Archimedean copulas. We also described three methods for fitting the model to data: maximum likelihood with the copula density, goodness-of-fit with the empirical copula-based Cramér von-Mises statistic and adversarial training by minimizing a divergence between true samples from data and fake samples from the copula. Empirically, the generative Archimedean copula was able to learn known copulas with different tail dependencies and fit real-world data. We also showed an extension to higher-dimensional data using hierarchical Archimedean copulas. Future work includes an end-to-end application such as pairs trading and architecture selection for the generative neural network.

\subsubsection*{Acknowledgements}
This work was supported in part by the Air Force Office of Scientific Research under award number FA9550-20-1-0397.

\bibliography{ng_241}

\onecolumn

\section{Supplementary Material}

\subsection{Probabilistic Construction of Archimedean and Hierarchical Archimedean Copulas}

Copulas can be derived from cumulative distribution functions (CDFs) via Sklar's theorem, i.e. specify a joint CDF $F$, compute univariate CDFs $F_1,\cdots,F_d$ from the joint CDF, then obtain the copula as $C(\mathbf{u})=F(F_1^{-1}(u_1),\cdots,F_d^{-1}(u_d)),\mathbf{u}\in[0,1]^d$. Sklar's theorem also applies to survival functions, i.e. for joint survival function $\bar{F}(\mathbf{x}) = P(X_1>x_1,\cdots,X_d>x_d), \mathbf{x}\in\mathbb{R}^d$, with univariate survival functions $\bar{F}_1,\cdots,\bar{F}_d$ where $\bar{F}_j=P(X_j>x_j)$, the copula which couples $\bar{F}$ to $\bar{F}_1,\cdots,\bar{F}_d$ is called the \emph{survival copula} and is given as the copula $C$ for which $\bar{F}(x) = C(\bar{F}_1(x_1),\cdots,\bar{F}_d(x_d))$.

\subsubsection{Archimedean Copulas}

We restate the probabilistic construction found in~\citep{joe2014dependence} Chapter 3.2, following~\citep{marshallolkin1988sampling}:

Let $G_1,\cdots,G_d$ be univariate CDFs. Let $Q\sim F_Q$ be a positive random variable with Laplace transform $\varphi_Q$, let $X_1,\cdots,X_d$ be dependent random variables that are conditionally independent given $Q=q$ such that $[X_j|Q=q]\sim G_j^q,\; q>0$.

The joint CDF is:
\begin{equation}
    F(x_1,\cdots,x_d)=\int_{0}^\infty G_1^q(x_1)\cdots G_d^q(x_d)dF_Q(q)=\varphi_Q(-\log G_1(x_1)-\cdots-\log G_d(x_d)),
\end{equation}
with univariate CDFs obtained from the joint CDF as:
\begin{equation}
    F_j(x_j)=\int_{0}^\infty G_j^q(x_j)dF_Q(q)=\varphi_Q(-\log G_j(x_j)), \;j\in\{1,\cdots,d\},
\end{equation}
and inverse:
\begin{equation}
    F_j^{-1}(u_j)=G_j^{-1}(\exp\{-\varphi_Q^{-1}(u_j)\}), \;u_j\in(0,1), \;j\in\{1,\cdots,d\},
\end{equation}
such that the copula via Sklar's theorem is:
\begin{equation}
    C(\mathbf{u})=F(F_1^{-1}(u_1),\cdots,F_d^{-1}(u_d))=\varphi_Q(\varphi_Q^{-1}(u_1)+\cdots+\varphi_Q^{-1}(u_d)), \; \mathbf{u}\in[0,1]^d.
\end{equation}

The multivariate extension of bivariate Archimedean copulas was introduced in~\citep{kimberling1974} with the condition that the above expression is a valid copula for any $d$ whenever $\varphi$, known as the \emph{generator} of the Archimedean copula, is \emph{completely monotone}, i.e. the Laplace transform of a positive random variable~\citep{bernstein1929fonctions, widder1941laplace}. The mixture representation with Laplace transform generators and an efficient algorithm for sampling from the mixture representation was subsequently given in~\citep{marshallolkin1988sampling}. 

We restate the sampling algorithm found in~\citep{mcneil2008sampling}, following~\citep{marshallolkin1988sampling}: 

Consider $(U_1,\cdots,U_d)=(\varphi(E_1/M),\cdots,\varphi(E_d/M))$, where $(E_1,\cdots,E_d)\sim i.i.d.\;\text{Exp}(1)$ are independent and identically distributed unit exponentials and $M\sim F_{M}$ is a positive random variable with Laplace transform $\varphi_M$.
\begin{equation}
\begin{split}
    P(U_1\leq u_1,\cdots,U_d\leq u_d)
    &=\int_0^\infty P(U_1\leq u_1,\cdots,U_d\leq u_d|M=s)dF_M(s)\\
    &=\int_0^\infty e^{-s(\varphi^{-1}(u_1)+\cdots+\varphi^{-1}(u_d)}dF_M(s)\\
    &=\varphi_M(\varphi_M^{-1}(u_1)+\cdots+\varphi_M^{-1}(u_d)).
\end{split}
\end{equation}

Thus an algorithm for sampling $\mathbf{U}\sim C$ is to sample $M\sim F_M$ with Laplace transform $\varphi_M$, sample $(E_1,\cdots,E_d)\sim i.i.d.\;\text{Exp}(1)$, then compute $(U_1,\cdots,U_d)=(\varphi_M(E_1/M),\cdots,\varphi_M(E_d/M))$.

\subsubsection{Hierarchical Archimedean Copulas}

A simple nested mixture representation involving Laplace transform generators was introduced in~\citep{joe1997multivariate}. Conditions for the nested copula to be a valid copula, called \emph{sufficient nesting conditions}, was derived based on the composition of an \emph{outer generator} $\varphi_0$ and an \emph{inner generator} $\varphi_1$ to get a completely monotone Laplace transform \emph{nested generator} $e^{-\nu_0\varphi_0^{-1}\circ\varphi_1}$, where $\nu_0$ is a positive random variable with Laplace transform $\varphi_0$, such that $\varphi_0,\varphi_1$ and $(\varphi_0^{-1}\circ\varphi_1)'$ are completely monotone. 

We illustrate with a simple three-dimensional example found in~\citep{mcneil2008sampling}, following~\citep{joe1997multivariate}. Consider the hierarchical Archimedean copula:
\begin{equation}
    C(u_1,u_2,u_3) = \varphi_0(\varphi_0^{-1}(u_1)+\varphi_0^{-1}\circ\varphi_1(\varphi_1^{-1}(u_2)+\varphi_1^{-1}(u_3))),
\end{equation}
where $\varphi_0,\varphi_1$ are Laplace transform generators of Archimedean copulas. We would like to express the above as a mixture of conditionally independent CDFs. Let $G_0$ be a distribution with Laplace transform $\varphi_0$:
\begin{align}
    C(u_1,u_2,u_3) 
    &= \varphi_0(\varphi_0^{-1}(u_1)+\varphi_0^{-1}\circ\varphi_1(\varphi_1^{-1}(u_2)+\varphi_1^{-1}(u_3)))\\\nonumber
    &= \int_0^\infty e^{-\nu_0\varphi_0^{-1}(u_1)}e^{-\nu_0\varphi_0^{-1}\circ\varphi_1(\varphi_1^{-1}(u_2)+\varphi_1^{-1}(u_3))}dG_0(\nu_0)\\ \nonumber
    &= \int_0^\infty F^{\nu_0}_0(u_1) C_{01}(F_0^{\nu_0}(u_2),F_0^{\nu_0}(u_3);\nu_0)dG_0(\nu_0),
\end{align}
where $F_0(\cdot):= e^{-\varphi_0^{-1}(\cdot)}$ and $F^{\nu_0}_0$ is a valid CDF for any $\nu_0>0$. In addition, $C_{01}(\cdot;\nu)$ is an Archimedean copula with Laplace transform generator $\varphi_{01}(\cdot;\nu_0)=e^{-{\nu_0}\varphi_0^{-1}\circ\varphi_1(\cdot)}$ and generator inverse $\varphi_{01}^{-1}(\cdot;\nu_0)=\varphi_1^{-1}\circ\varphi_0(-\log(\cdot)/\nu_0)$, such that $C_{01}(\cdot;\nu_0)$ taking marginals $F_0^{\nu_0}(u_2)$ and $F_0^{\nu_0}(u_3)$ as inputs gives:
\begin{align}
    C_{01}(F_0^{\nu_0}(u_2),F_0^{\nu_0}(u_3);\nu_0) 
    &=  \varphi_{01}(\varphi_{01}^{-1}(F_0^{\nu_0}(u_2);\nu_0)+\varphi_{01}^{-1}(F_0^{\nu_0}(u_3);\nu_0);\nu_0) \\\nonumber
    & = e^{-{\nu_0}\varphi_0^{-1}\circ\varphi_1( \varphi_1^{-1}\circ\varphi_0(-\log(e^{-{\nu_0}\varphi_0^{-1}(u_2)})/{\nu_0}) + \varphi_1^{-1}\circ\varphi_0(-\log(e^{-{\nu_0}\varphi_0^{-1}(u_3)})/{\nu_0})  )}\\\nonumber
    & = e^{-{\nu_0}\varphi_0^{-1}\circ\varphi_1(\varphi_1^{-1}(u_2)+\varphi_1^{-1}(u_3))}.
\end{align}
The completely monotone property of the Laplace transform generator $\varphi_{01}(\cdot;\nu)=e^{-\nu\varphi_0^{-1}\circ\varphi_1(\cdot)}$ then implies $(\varphi_0^{-1}\circ\varphi_1)'$ is completely monotone. In addition, letting $G_{01}(\cdot;\nu_0)$ be a distribution with Laplace transform $\varphi_{01}(\cdot;\nu_0)$, we express the hierarchical Archimedean copula as a nested mixture of conditionally independent CDFs:
\begin{align}
    C(u_1,u_2,u_3) 
    &= \int_0^\infty F^{\nu_0}_0(u_1) C_{01}(F_0^{\nu_0}(u_2),F_0^{\nu_0}(u_3);\nu_0)dG_0(\nu_0)\\\nonumber
    &= \int_0^\infty F^{\nu_0}_0(u_1) \int_0^\infty e^{-{\nu_{01}}\varphi_{01}^{-1}(F_0^{\nu_0}(u_2))}e^{-{\nu_{01}}\varphi_{01}^{-1}(F_0^{\nu_0}(u_3))}dG_{01}({\nu_{01}};\nu_0)dG_0(\nu_0)\\\nonumber
    &= \int_0^\infty F^{\nu_0}_0(u_1) \int_0^\infty e^{-{\nu_{01}}\varphi_1^{-1}(u_2)}e^{-{\nu_{01}}\varphi_1^{-1}(u_3)}dG_{01}({\nu_{01}};\nu_0)dG_0(\nu_0)\\\nonumber
    &= \int_0^\infty \int_0^\infty F^{\nu_0}_0(u_1) F^{\nu_{01}}_1(u_2)F^{\nu_{01}}_1(u_3)dG_{01}({\nu_{01}};\nu_0)dG_0(\nu_0),\\\nonumber
\end{align}
where $F_1(\cdot):= e^{-\varphi_1^{-1}(\cdot)}$ and $F^{\nu_{01}}_1$ is a valid CDF for any $\nu_{01}>0$.

This construction and condition were restated for nesting to arbitrary depth in~\citep{mcneil2008sampling}. 

Based on the mixture representation, \citet{mcneil2008sampling} also provided algorithms for sampling nested Clayton and nested Gumbel copulas. It was also showed that Clayton and Gumbel copulas are unfortunately not compatible for nesting. The challenge was to find combinations of known distributions with $\varphi_0,\varphi_1$ and $e^{-\nu_0\varphi_0^{-1}\circ\varphi_1}$ as their Laplace transforms. Sampling using~\citet{mcneil2008sampling}'s algorithm for nested Ali-Mikhail-Haq, nested Frank, nested Joe, more parametric families and numerical inversion of Laplace transform was by~\citep{hofert2008sampling}. It was subsequently recognized in~\citep{hering2010_levy} that the sufficient nesting condition for $(\varphi_0^{-1}\circ\varphi_1)'$ to be completely monotone can be satisfied by letting $\varphi_1=\varphi_0\circ\psi_1$, where $\psi_1$ is the Laplace exponent, with completely monotone derivative, in the Laplace transform of Lévy subordinators.

We restate the probabilistic construction with Lévy subordinators from~\citep{hering2010_levy}:

For each `time' $t\geq0$, the Laplace transform of a Lévy subordinator $\Lambda_t$, i.e. a non-decreasing Lévy processes such as the compound Poisson process,  is given as:
\begin{equation}
    \mathbf{E}[e^{-x\Lambda_t}]=e^{-t\psi(x)},
\end{equation}
where $\psi(x)$ is the Laplace exponent.

Consider $(\frac{E_{1,1}}{\Lambda_1(M)},\cdots,\frac{E_{1,d_1}}{\Lambda_1(M)},\cdots,\cdots,\cdots,\frac{E_{J,1}}{\Lambda_J(M)},\cdots,\frac{E_{J,d_J}}{\Lambda_J(M)})$, where $E_{j,i}\sim i.i.d.\;\text{Exp}(1)$, $\Lambda_j$ are Lévy subordinators with Laplace exponents $\psi_j$, and $\Lambda_j$ are evaluated at a common `time' $t=M$, where M is a positive random variable with Laplace transform $\varphi_0$. The hierarchical Archimedean copula is then constructed using the survival analog of Sklar's theorem.

The joint survival function is:
\begin{equation}
\begin{split}
    P(\frac{E_{j,i}}{\Lambda_j(M)}>x_{j,i}\text{\;for all}j,i)
    &=\mathbf{E}[\prod_{j=1}^{J}\prod_{i=1}^{d_J}e^{-\Lambda_j(M)x_{j,i}}]\\
    &=\mathbf{E}[\prod_{j=1}^{J}e^{-\Lambda_j(M)\sum_{i=1}^{d_J}x_{j,i}}]\\
    &=\mathbf{E}[\prod_{j=1}^Je^{-M\psi_j(\sum_{i=1}^{d_j}x_{j,i})}]\\
    &=\mathbf{E}[e^{-M\sum_{j=1}^J\psi_j(\sum_{i=1}^{d_j}x_{j,i})}]\\
    &=\varphi_0(\sum_{j=1}^J\varphi_0^{-1}\circ(\varphi_0\circ\psi_j)(\sum_{i=1}^{d_j}x_{j,i})),
\end{split}
\end{equation}
and each component $\frac{E_{j,i}}{\Lambda_j(M)}$ has survival function:
\begin{equation}
    P(\frac{E_{j,i}}{\Lambda_j(M)}>x) =\mathbf{E}[e^{-x\Lambda_j(M)}] =\mathbf{E}[e^{-M\psi_j(x)}] =(\varphi_0\circ\psi_j)(x).
\end{equation}

Using the survival analog of Sklar's theorem, given the above univariate survival functions, the hierarchical Archimedean copula $C$ with outer generator $\varphi_0$ and inner generators $\varphi_j=\varphi_0\circ\psi_j$, we recover the above joint survival function.

\subsection{Experiment Details}

Following the experiment setup in~\citep{ling2020_ACNet}, the commonly-used copulas were the Clayton, Frank and Joe copulas, each governed by a single parameter and chosen to be 5, 15, and 3 respectively. Each dataset had 2000 train and 1000 test points. The real-world data were the Boston housing, Intel-Microsoft (INTC-MSFT) stocks and Google-Facebook (GOOG-FB) stocks. Each dataset was divided into train and test points in a 3:1 ratio, then rank-normalized to get approximately uniform margins. 

Similar to the experiment parameters in~\citep{ling2020_ACNet}, the tolerance for Newton's root-finding method was 1e-10. The generative neural network was a multilayer perceptron of comparable size, 2 hidden layers, each of width 10. We used $\text{U}(0,1)$ as the input source of randomness, default weight initialization, LeakyReLU intermediate activations and $\exp(.)$ output activation. For training with maximum likelihood, we used the same optimization parameters: stochastic gradient descent (SGD) with learning rate 1e-5 and momentum 0.9 on sum of log-likelihoods. For training with goodness-of-fit, we used SGD with learning rate 1e-3 and momentum 0.9. For adversarial training, we used Adam with learning rate 1e-4, momentum 0.9 and betas (0.5, 0.999). The discriminative neural network had a single hidden layer of width 20, default weight initialization, LeakyReLU intermediate activations and $\text{sigmoid}(.)$ output activation. All training methods used the same batch size of 200 and converged within 10k epochs. We reported the results at 10k epoch. Experiments were conducted using PyTorch, on a 2.7 GHz Intel Core i7 with 16 GB of RAM.

To reduce computation complexity during training, we used a smaller number $L=100$ samples from the generative neural network to approximate the Laplace transforms. To increase inference accuracy for evaluation, we used a larger number $L=1000$ samples from the generative neural network to approximate the Laplace transforms. 

\subsubsection{Enforcing Structural Properties}

\begin{figure}[!h]
    \centering
    \includegraphics[width=0.5\linewidth]{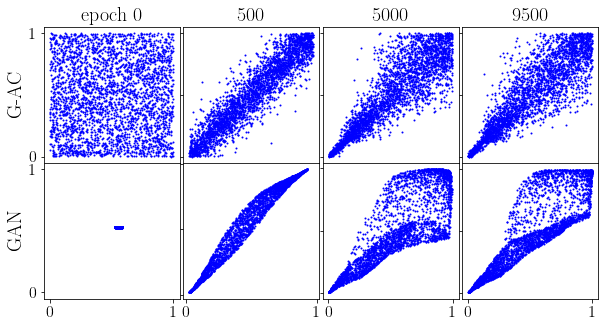}
    \caption{Training progression at epochs 0, 500, 5000 and 10000, for learning a Clayton copula. Samples from our copula, shown on top, must satisfy an Archimedean copula, while that from a vanilla GAN, shown below, may not. }\label{fig:samples_training_GAN}
\end{figure}

Compared to vanilla GAN~\citep{goodfellow2014gan}, our generating network must satisfy an Archimedean copula. We show this via the training progression for learning a Clayton copula in~\figurename~\ref{fig:samples_training_GAN}.

\end{document}